\documentclass[10pt, a4paper]{article}

\usepackage[final]{lrec2026} 
\usepackage{graphicx}
\usepackage{todonotes}
\usepackage{booktabs}
\usepackage{comment}
\usepackage{amsmath}
\usepackage{array}
\usepackage{subcaption}
\usepackage{colortbl}
\usepackage{xcolor}
\usepackage{tikz}
\usepackage{listings}
\usepackage[ruled,vlined]{algorithm2e}
\usepackage{url}
\usepackage{seqsplit}
\usepackage{siunitx}  
\usepackage{makecell}
\usepackage{enumitem}
\usepackage{amssymb} 
\newcommand{\parent}{\ensuremath{\ \hookrightarrow\ }}
\newcommand{\smalltag}[1]{\texttt{\small #1}}
\usepackage{xcolor}
\usepackage[most]{tcolorbox}
\usepackage[ruled,vlined]{algorithm2e}
\tcbset{
    promptstyle/.style={
        enhanced,
        width=\linewidth,
        colback=white,
        colframe=black,
        colbacktitle=gray!20,
        coltitle=black,
        rounded corners,
        boxrule=0.5pt,
        drop shadow=black!50!white,
        attach boxed title to top left={
            xshift=-2mm,
            yshift=-2mm
        },
        boxed title style={
            rounded corners,
            size=small,
            colback=gray!20
        }
    },
    replystyleg/.style={
        enhanced,
        width=\linewidth,
        colback=green!15,
        colframe=black,
        colbacktitle=green!30,
        coltitle=black,
        boxrule=0.5pt,
        drop shadow=black!50!white,
        rounded corners,
        sharp corners=north,
        attach boxed title to top right={
            xshift=-2mm,
            yshift=-2mm
        },
        boxed title style={
            rounded corners,
            size=small,
            colback=green!40
        }
    },
    replystyler/.style={
        enhanced,
        width=\linewidth,
        colback=red!15,
        colframe=black,
        colbacktitle=red!40,
        coltitle=black,
        boxrule=0.5pt,
        drop shadow=black!50!white,
        rounded corners,
        sharp corners=north,
        attach boxed title to top right={
            xshift=-2mm,
            yshift=-2mm
        },
        boxed title style={
            rounded corners,
            size=small,
            colback=red!40
        }
    }
}

\newtcolorbox{promptbox}[1][]{
    promptstyle,
    title=Prompt,
    #1
}

\lstdefinelanguage{json}{
    basicstyle=\small\ttfamily,
    stepnumber=1,
    numbersep=8pt,
    xleftmargin=15pt,
    showstringspaces=false,
    breaklines=true,
    frame=lines,
    backgroundcolor=\color{gray!10},
    literate=
     *{0}{{{\color{blue}0}}}{1}
      {1}{{{\color{blue}1}}}{1}
      {2}{{{\color{blue}2}}}{1}
      {3}{{{\color{blue}3}}}{1}
      {4}{{{\color{blue}4}}}{1}
      {5}{{{\color{blue}5}}}{1}
      {6}{{{\color{blue}6}}}{1}
      {7}{{{\color{blue}7}}}{1}
      {8}{{{\color{blue}8}}}{1}
      {9}{{{\color{blue}9}}}{1}
      {:}{{{\color{red}{:}}}}{1}
      {,}{{{\color{red}{,}}}}{1}
}
\definecolor{codegreen}{rgb}{0,0.6,0}
\definecolor{codegray}{rgb}{0.5,0.5,0.5}
\definecolor{ForestGreen}{RGB}{34, 139, 34}

\definecolor{backcolour}{rgb}{0.95,0.95,0.92}
\lstdefinestyle{jsonstyle}{
    backgroundcolor=\color{backcolour},
    commentstyle=\color{codegreen},
    keywordstyle=\color{ForestGreen},
    numberstyle=\tiny\color{limegreen},
    stringstyle=\color{limegreen},
    basicstyle=\small\ttfamily,
    breaklines=true,
    keepspaces=true,
    numbers=none,
    showstringspaces=false,
    tabsize=2,
    morekeywords={label, start_date_for_period, end_date_for_period, currency_unit, value},
    moredelim=**[is][\color{red}\bfseries]{@@}{@@} 
}

\makeatletter
\newcommand{\printfontsize}{Font size is: \f@size pt}
\makeatother

\def\checkmark{\tikz\fill[scale=0.4](0,.35) -- (.25,0) -- (1,.7) -- (.25,.15) -- cycle;}
\usepackage{todonotes}
\newcommand{\hifi}{\textsc{HiFi-KPI}}

\title{\textsc{HiFi-KPI}:\\
A Dataset for Hierarchical KPI Extraction from Earnings Filings}

\name{\begin{tabular}{c}
    Rasmus T. Aavang$^{1,2}$, Giovanni Rizzi$^2$, Rasmus Bøggild$^2$ \\
    Alexandre Iolov$^2$, Mike Zhang$^{3,1,4}$, Johannes Bjerva$^1$
\end{tabular}}

\address{$^1$Department of Computer Science, Aalborg University, Denmark \\
         $^2$ALIPES ApS, Denmark \\
         $^3$University of Copenhagen, Denmark \\
         $^4$Pioneer Centre for AI, Denmark \\
         \texttt{rtaj@cs.aau.dk}\\}

\abstract{
Accurate tagging of earnings reports can yield significant short-term returns for stakeholders. 
The machine-readable inline eXtensible Business Reporting Language (iXBRL) is mandated for public financial filings. 
Yet, its complex, fine-grained taxonomy limits the cross-company transferability of tagged Key Performance Indicators (KPIs). 
To address this, we introduce the \textbf{Hi}erarchical \textbf{Fi}nancial \textbf{K}ey \textbf{P}erformance \textbf{I}ndicator (\hifi{}) dataset, a large-scale corpus of 1.65M paragraphs and 198k unique, hierarchically organized labels linked to iXBRL taxonomies. 
\hifi{} supports multiple tasks and we evaluate three: KPI classification, KPI extraction, and structured KPI extraction. 
For rapid evaluation, we also release \textbf{\hifi{}-Lite}, a manually curated 8K paragraph subset. Baselines on \hifi{}-Lite show that encoder-based models achieve over 0.906 macro-F1 on classification, while Large Language Models (LLMs) reach 0.440 F1 on structured extraction. 
Finally, a qualitative analysis reveals that extraction errors primarily relate to dates. We open-source all code and data at \url{https://github.com/aaunlp/HiFi-KPI}.}

\begin{document}

\maketitleabstract

\section{Introduction}
\definecolor{lightblue}{RGB}{43, 123, 170}
\definecolor{tomato}{RGB}{199, 93, 93}
\definecolor{navyblue}{RGB}{107, 152, 147}
\definecolor{amber}{RGB}{230, 175, 46}
\definecolor{limegreen}{RGB}{126, 217, 87}
\definecolor{context}{RGB}{173, 235, 104}
\definecolor{context_dark1}{RGB}{138, 188, 83}
\definecolor{context_dark2}{RGB}{104, 141, 62}

\definecolor{Calculation}{RGB}{55, 156, 151}
\definecolor{presentationLayerColor}{RGB}{208, 149, 149}

\definecolor{purpleish}{RGB}{195, 49, 255}

\begin{figure}[t]
    \centering
    \includegraphics[width=\linewidth, trim=0 2.5cm 0 0, clip]{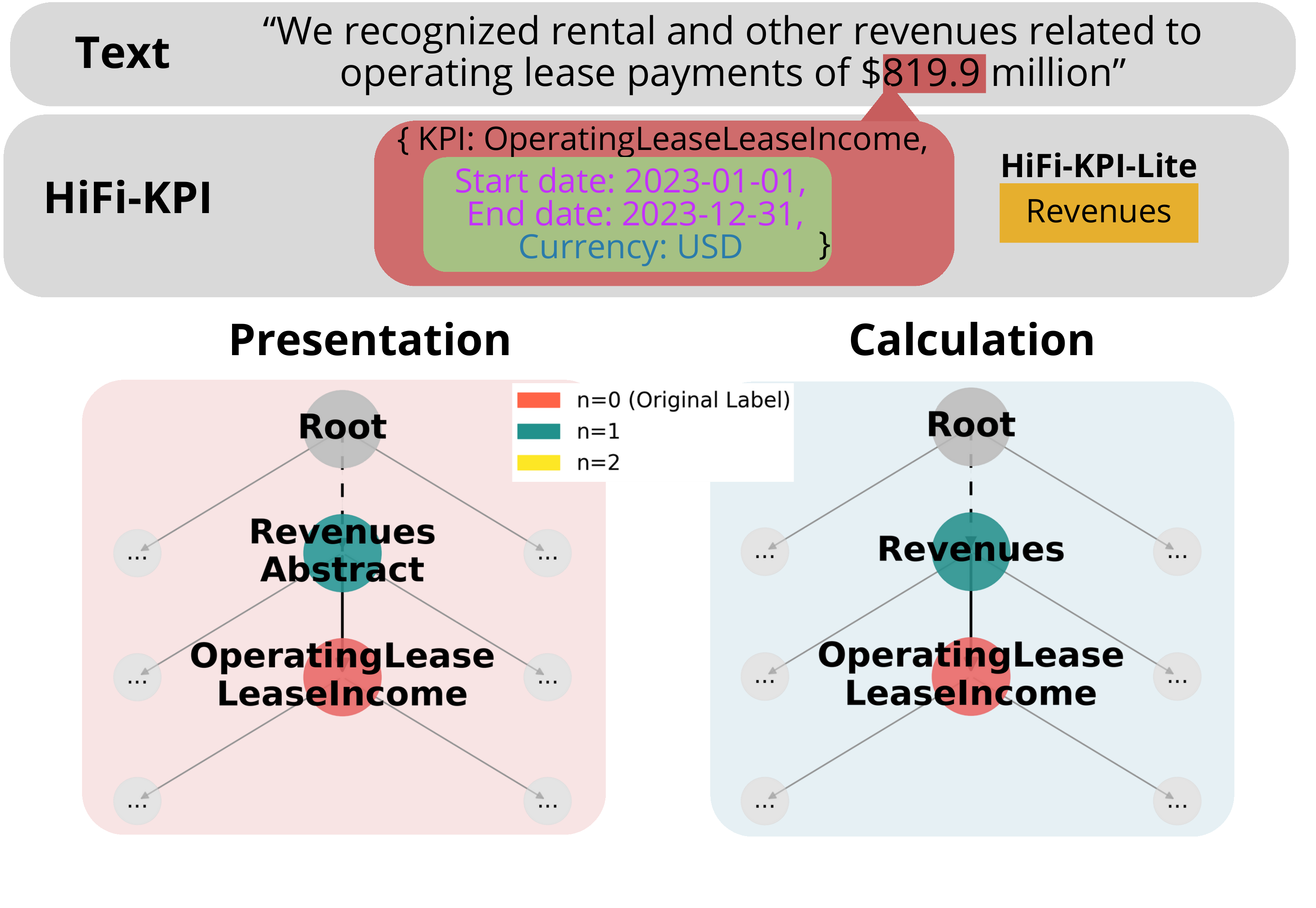}
    \caption{\textbf{\hifi{}} provides valuable \textcolor{context_dark1}{contextual information} for key performance indicators, including associated \textcolor{magenta}{time periods} and \textcolor{lightblue}{currencies}. 
    Additionally, \hifi{} offers hierarchical context, demonstrated by the hyper-specific \textcolor{tomato}{OperatingLeaseLeaseIncome} mapping to \textcolor{navyblue}{RevenuesAbstract (n=1)} in the \textcolor{presentationLayerColor}{Presentation taxonomy}, and \textcolor{navyblue}{Revenues (n=1)} in the \textcolor{Calculation}{Calculation taxonomy}.
    Finally, \textbf{\hifi{}-Lite} provides expert mapping to \textcolor{amber}{Revenues}.
        }
    \label{fig:figure1}
\end{figure}

As unstructured data in finance grows~\cite{doi:10.1080/00014788.2019.1611730}, there is a strong interest in developing NLP benchmarks and methods.
Financial reporting standards in the EU, UK, USA, and others require public companies to file financial summaries with iXBRL. 
iXBRL provides machine-readable tags for key information.
However, these tags are highly fine-grained (up to 198K labels), limiting generalization (see Figure \ref{fig:figure1}).
Making annotation difficult, time-consuming, and expensive, even for expert taggers, with over 34\% of documents containing errors~\cite{Bricker2020XBRL}.
Although several datasets have been created for the financial domain~\cite{chen2022convfinqa, jorgensen-etal-2023-multifin} and even for financial reports~\cite{loukas2021edgar, loukas-etal-2022-finer, sharma-etal-2023-financial, lai2024sec}, the rich structural information embedded by iXBRL remains an untapped resource for enhancing contextual understanding 
and enabling cross-company generalization. 

We present the first resource that provides context for the KPI tags present in SEC filings.
iXBRL contains several taxonomies. 
We focus on two: the Presentation taxonomy that dictates the layout in reports and the Calculation taxonomy that defines arithmetic relationships.
We provide context for the KPI tags by creating a unified Presentation taxonomy across companies, and the same for the Calculation taxonomy.
Further, we are the first to provide KPI tag context by preserving the relationships with time periods, numerical values, and currencies.
Our research into leveraging this rich structure is guided by the following questions:

\textbf{RQ1.} To what extent can the structural hierarchies of iXBRL taxonomies be leveraged to generalize hyper-specific financial labels for automated information extraction?

\textbf{RQ1.1.} Which iXBRL taxonomy, Presentation or Calculation, provides the most effective structural basis, and how does model performance vary across different levels of label granularity?

\textbf{RQ1.2.} How well do SOTA Large Language Models extract information from expert-defined labels?

Accurate information extraction from iXBRL has great value as accurate analysis of earnings reports can yield significant short-term returns for investors \citep{ke2005institutional}, besides helping corporate compliance.
We investigate this by contributing:

\begin{figure*}[t]
    \centering
    \includegraphics[trim=0 800bp 0 0, clip, width=\linewidth]{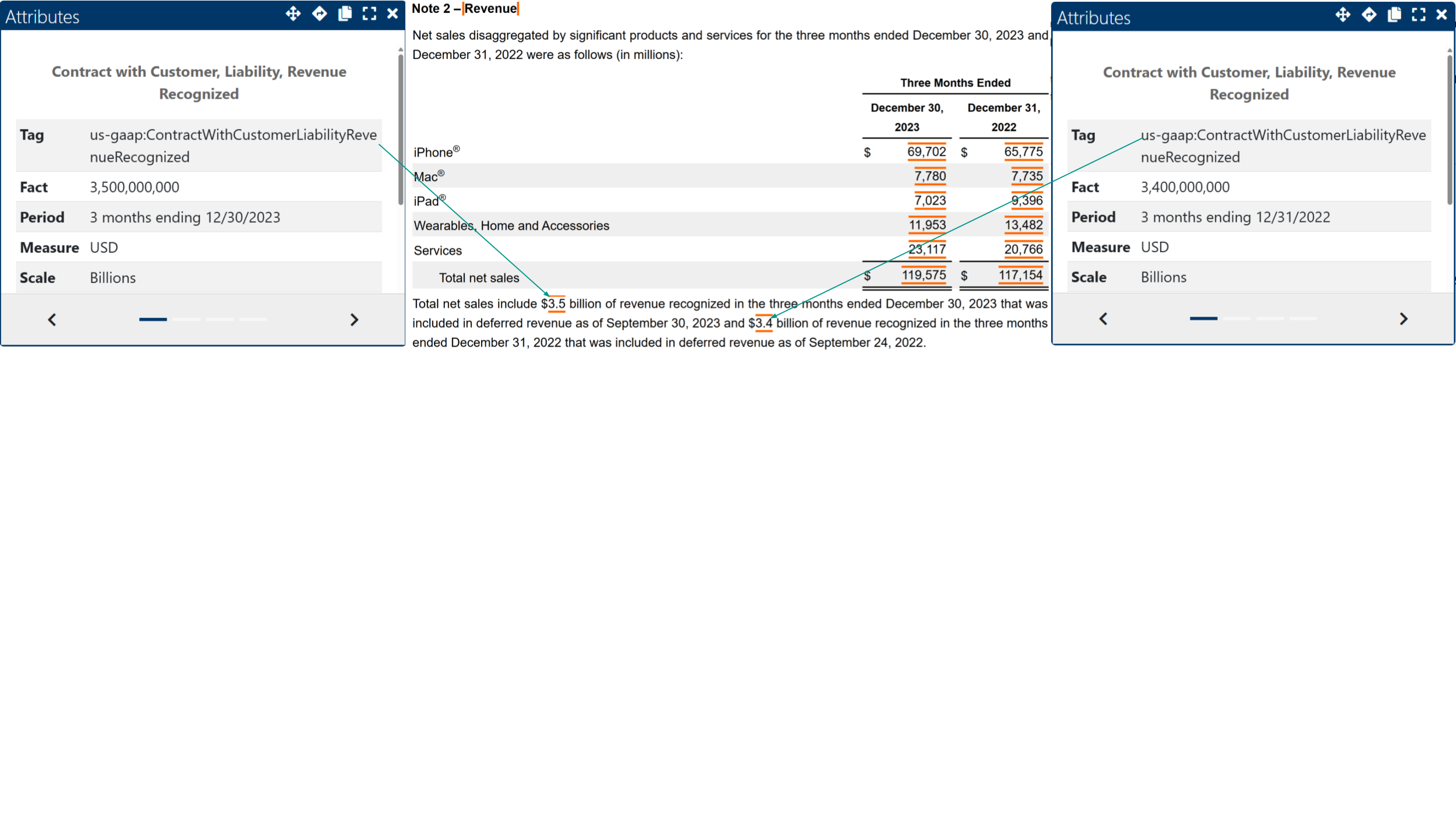}
    \caption{\textbf{Example \hifi{}.} The dataset is created by scraping the contextual snippets often following tables in SEC filings; Example is from the SEC's Inline XBRL Viewer for the 10Q for AAPL Q4 2023.}
    \label{fig:sec-filling}
\end{figure*}
\begin{itemize}
    \itemsep0em
    \item \hifi{}: An iXBRL-based dataset with 1.65M paragraphs and 4.5M entities, and a smaller \hifi{}-Lite with four expert-defined labels.
    \item Two cleaned and unified taxonomies enabling user-specified levels of granularity with our granularity selection method that overcomes the hyper-specificity of the original labels. 
    \item Baselines for text classification, sequence labeling, and LLM-based structured extraction for \hifi{}-Lite.
\end{itemize}

\section{Related Work}
\begin{table}[t]
\centering
\scriptsize
\setlength{\tabcolsep}{4pt}
\begin{tabular}{lrrcc}
\toprule
\textbf{Resource} & \textbf{Samples} & \thead{\scriptsize \textbf{Unique} \\ \scriptsize \textbf{Labels}} & \textbf{Context} & \textbf{Taxonomy} \\
\midrule
\citet{loukas-etal-2022-finer} & 1.12M & 139 & \textcolor{red}{\text{$\boldsymbol{\times}$}} & \textcolor{red}{\text{$\boldsymbol{\times}$}} \\
\citet{sharma-etal-2023-financial} & 79K & 2.8K & \textcolor{red}{\text{$\boldsymbol{\times}$}} & \textcolor{red}{\text{$\boldsymbol{\times}$}} \\
\hifi{} (Ours) & 1.65M & 198K & \textcolor{limegreen}{\checkmark} & \textcolor{limegreen}{\checkmark} \\
\hifi{}-Lite (Ours) & 8K & 5 & \textcolor{limegreen}{\checkmark} & N/A  \\
\bottomrule
\end{tabular}
\caption{Comparison of \hifi{} \& \hifi{}-Lite and previous SEC filings based resources.}
\label{tab:resources}
\end{table}

\paragraph{Financial NLP Datasets.}
Financial NLP datasets cover sentiment analysis~\citep{gupta2020comprehensive}, named entity recognition~\citep{alvarado2015domain, shah-etal-2022-flue}, and numerical reasoning~\citep{chen2022finqadatasetnumericalreasoning}.
While early approaches used rule-based methods~\citep{extraction2007, sheikh2012rule, hutto2014vader}, modern efforts leverage large corpora e.g. from SEC Filings ~\citep{loukas2021edgar}. 
KPI-EDGAR~\citep{10069806} that defines an NER and relation extraction task. 
Additionally, FiNER-139~\citep{loukas-etal-2022-finer} and \citet{sharma-etal-2023-financial}, which both leverage SEC filings to create a sequence labeling task with a limited label set.
FiNER-139 limiting itself to the 139 most common labels, resulting in high sparsity (80.42\% untagged entries) and \citet{sharma-etal-2023-financial} using US-GAAP metrics only; 
\hifi{} introduces valuable context, enabling more complex generative tasks and taxonomies, which in turn facilitate the analysis of the full label set. 
This makes \hifi{} the first resource to utilize the full labelset, resulting in 0\% untagged entries, and a density of 2.77 tags/entry.

\paragraph{Language Models in Finance.}
Transformer-based models~\citep{vaswani2017attention} has been a huge influence on the NLP field, leading to several specialized models for the finance domain, a notable example is the proprietary LLM BloombergGPT \cite{wu2023bloomberggpt}, and commonly used open-source financial language models are FinBERT variants \citep{yang2020finbert, araci2019finbertfinancialsentimentanalysis}. To move from sentiment analysis to granular data extraction, we focus on sequence labeling. 
While pretrained models like SEC-BERT~\citep{loukas-etal-2022-finer} exist, a fine-tuned token classification version is not public. 
We therefore use google-bert/bert-base-uncased~\cite{devlin2019bertpretrainingdeepbidirectional} to establish a reproducible baseline.
Sentence transformers~\citep{reimers2019sentencebertsentenceembeddingsusing} create semantically informative sentence embeddings orders of magnitude faster than BERT, making them ideal for quickly creating static representations to be able to iterate over many different label spaces for our hierarchical task.
The introduction of GPT-3~\cite{brown2020languagemodelsfewshotlearners} popularized decoder-only architectures. 
This new paradigme lead to state-of-the-art open-source models such as gemma-3-27B~\cite{gemmateam2025gemma3technicalreport}, DeepSeek-V3.1~\cite{deepseekai2024deepseekv3technicalreport}, Qwen3-30B-A3B~\citep{qwen3technicalreport} and mistral-Small-3.2-24B~\citep{mistral2025small}.
\hifi{} can benchmark these generative models not only on extracting the correct label but also on extracting important contextual information for these labels.    

\paragraph{SEC Filings}
U.S. public companies must file quarterly (10-Q) and annual (10-K) reports~\citep{SECFinalRule2000,sec_exchange_2024}, which contain standardized financial statements. 
Since June 15, 2020, SEC filings follow the inline eXtensible Business Reporting Language (iXBRL) open standard for accounting data.\cite{sec2018, intrinio_xbrl}. 
iXBRL is a standard that enables automatic computational extraction of tagged KPIs.
SEC filings in iXBRL contain detailed financial statements that often follow standard templates. (see Figure \ref{fig:sec-filling})
\hifi{} contains the contextual data available in iXBRL.
This context is paramount for actually assessing companies' financial situation, as one will often see KPIs reported in relation to the previous quarter or year.
A common formulation is like the following
\begin{quote}
\$0.52 and \$0.34 per basic share, or \$0.49 and \$0.34 per diluted share, respectively, for the three months ended March 31, 2020 and 2019, respectively. 
\end{quote}
The word "respectively" appears over 650k times in the dataset, highlighting the need to link figures like "earnings per share" to their correct time period.
\begin{quote}
The 2019 facilities consist (...) (i) a \$675.0 million United States dollar-denominated revolving credit facility, (ii) a CAD  \$70.0 million Canadian dollar-denominated (..) (iii) a €200.0 million euro-denominated ...
\end{quote}
Since the data includes figures in USD, EUR, and CAD, conflating these currencies would be a fundamental error due to their vastly different value.

\paragraph{iXBRL Taxonomies}
iXBRL is highly complex, just the US-GAAP (Generally Accepted Accounting Principles) taxonomy, that describes the accounting standards of the U.S. Securities and Exchange Commission (SEC), contains about 17,000 unique tags~\cite{intrinio_xbrl}. 
iXBRL guidelines require annotators to choose the most specific tag.
Lastly, each company can also extend the taxonomy.
This leads to companies using iXBRL differently, with highly specific and company-based tags. e.g.
\begin{itemize}
    \item \smalltag{\seqsplit{aapl:EquitySecuritiesFVNIAccumulatedGrossUnrealizedGainBeforeTax}}
    \item \smalltag{\seqsplit{tsla:LeasedAssetsNet}}
\end{itemize}
Because company-specific tags are not standardized across firms, iXBRL provides a framework of relations and taxonomies to aid comparisons.
This paper focuses on two of them as they define key financial meaning 
\begin{description}[font=\bfseries, style=unboxed, itemsep=0.5em]
    \item[\smalltag{.cal} (Calculation)]
    Describes the arithmetic relations between concepts. 
     (e.g., “fact A + fact B = fact C”)~\cite{XBRLTaxonomies}. 
    
    {For example,} Tesla’s tag direct parent, \smalltag{tsla:LeasedAssetsNet} \parent \smalltag{us-gaap:Assets}. for Apple, we see \smalltag{aapl:Equity...GainBeforeTax} \parent \smalltag{us-gaap:EquitySecuritiesFvNiCost}, both aggregating to \smalltag{us-gaap:AssetsNet} at the root.

    \item[\smalltag{.pre} (Presentation)]
    Which arranges tags in a structure that is appropriate to represent the hierarchical relationships in business data. \citep{openriskmanual_xbrl, XBRL_Presentation}.
    
    {For example,} Tesla's tag has a direct parent: \smalltag{tsla:LeasedAssetsNet} \parent \smalltag{us-gaap:AssetsNoncurrentAbstract}. 
    Apple's tag have a different direct parent, but both share the parent \smalltag{\seqsplit{us-gaap:StatementOfFinancialPositionAbstract}}.
\end{description}
\hifi{} enables viewing these company-specific tags, as represented by their parents in the taxonomy, at a desired granularity.
We release the code for the implementation of our bottom-up granularity selection algorithm for easy use.

\section{\hifi{} }

\begin{table}[t]
\centering
\resizebox{\columnwidth}{!}{
\begin{tabular}{lllllll}
\toprule
 & \multicolumn{2}{c}{\textbf{Train (Lite)}} & \multicolumn{2}{c}{\textbf{Dev. (Lite)}} & \multicolumn{2}{c}{\textbf{Test (Lite)}} \\
\midrule
\textbf{Cutoff Date} & \multicolumn{2}{c}{2023.10.31} & \multicolumn{2}{c}{2024.05.31} & \multicolumn{2}{c}{2024.06.01} \\
\textbf{\# Paragraphs} & 1.33M & (6,194) & 149K & (755) & 168K & (847) \\
\textbf{\# Entities} & 3.64M & (19,043) & 418K & (2,545) & 448K & (2,399) \\
\textbf{Avg. Entities} & 2.73 & (3.07) & 2.80 & (3.37) & 2.67 & (2.83) \\
\textbf{Avg. Words} & 87.31 & (80.10) & 87.69 & (83.08) & 86.26 & (77.20) \\
\textbf{Avg. Length (Chars)} & 556.00 & (514.85) & 558.46 & (538.65) & 549.68 & (498.06) \\
\bottomrule
\end{tabular}
}
\caption{\textbf{Dataset Statistics.} We show the full dataset and the lite version statistics in brackets.}
\label{tab:merged_dataset_statistics}
\end{table}

\begin{figure}[t]
    \centering
    \includegraphics[width=\linewidth]{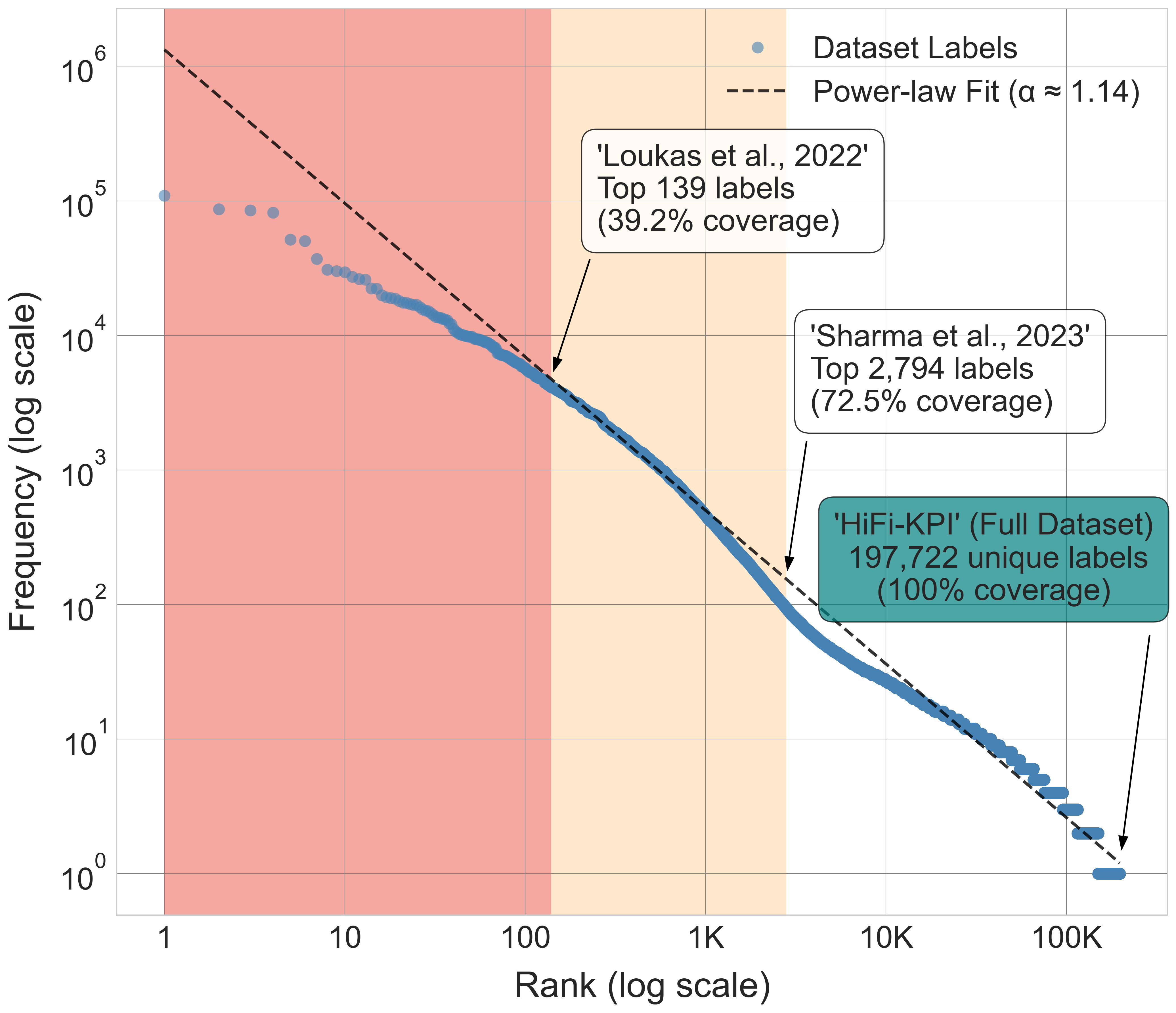}
    \caption{\textbf{Rank-frequency Distribution.} iXBRL labels on a log-log scale follow a power-law distribution, indicated by the fitted dashed line. Shaded regions highlight the cumulative coverage of all labels in \hifi{} by the most frequent label count from \citep{loukas-etal-2022-finer} and \citep{sharma-etal-2023-financial}}
    \label{fig:rank-size}
\end{figure}

\hifi{} supports multiple downstream tasks such as text classification, sequence labeling, structured information extraction, multi-label classification, and financial question answering. Table~\ref{tab:merged_dataset_statistics} summarizes key statistics for both the full dataset and the lite subset with expert labels made for fast inference.  
We collected all quarterly and annual reports published between 2017-01-01 and 2024-06-01, yielding 42,018 quarterly and 14,389 annual reports. 
We parsed each iXBRL document using \texttt{beautifulsoup} \citep{richardson2007beautiful} and regular expressions to extract text spans along with every iXBRL tag, associated time period, and numeric values. 
Any unparsed spans were discarded.
The extracted iXBRL tags show high data integrity with no snippets that wrongly include a preceding dollar sign ($\$$) or a succeeding magnitude specifier.
From these steps, we obtained 1.9M paragraphs (1.11M from quarterly reports; 0.74M from annual) and 5.3M tagged entities.
The long-tailed label distribution (Figure \ref{fig:rank-size}) approximates a power law, though the most frequent labels occur even less often than suggested by the fit.

\paragraph{Dataset Refinement and Quality Evaluation}
We evaluate the original company annotators' accuracy as well as the parsed entries' quality by manual verification of samples. 
We randomly sampled 100 entries from each data split (train, validation, test).
We mitigate annotator fatigue by presenting half of the entries in the order of train, validation, and test, while the remaining half were presented in the reverse order. 
We ask an industry professional to locate errors with respect to character span (Span), temporal period (Date), currency (Currency), and value (Value). 
Further, we asked the annotator to note which entries seemed to be incorrectly parsed. This could be that the start or end of the text snippet suggests that there is more relevant context. 
Since the iXBRL standard is highly complex, we were not able to assess the quality of the iXBRL tag chosen by the company annotator.
The results of this test can be seen in Table \ref{tab:error_analysis_comparison}.
Although the test revealed a low overall error rate, we used the reported errors to further refine the dataset, creating regular expressions to filter out snippets with ``false starts'' (e.g., those beginning with whitespace or a non-capitalized letter).
After filtering out the errors, we ran the same experiment again with 100 newly sampled snippets from the train, validation, and test splits for a total of 600 checked samples. We find under 1\% of entities having any errors, resulting in the final cleaned dataset having 1.65M snippets and 4.5M entities.

\begin{figure}[t]
    \centering
    \includegraphics[width=\linewidth]{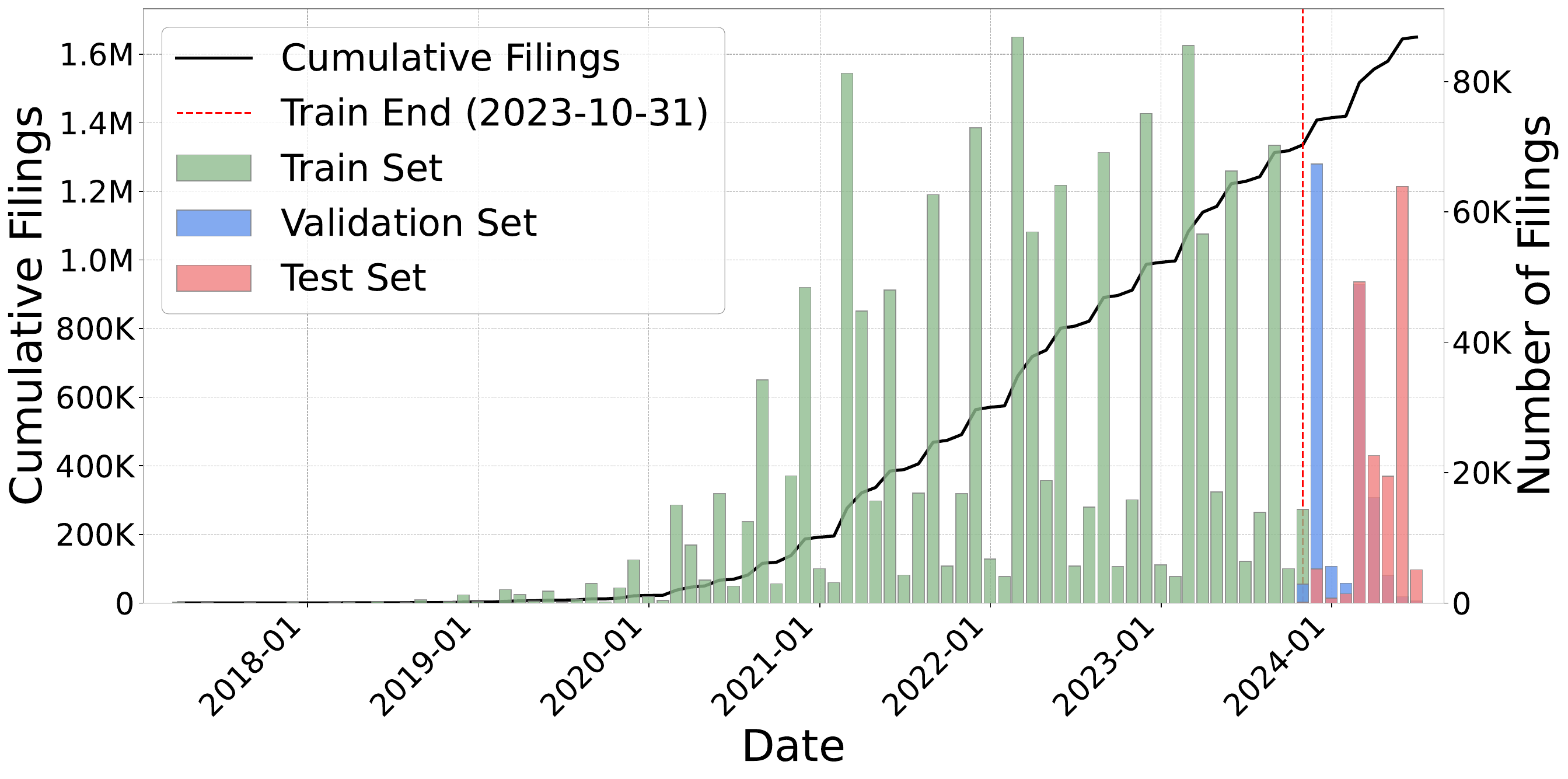}
    \caption{\textbf{Temporal Data Split} \hifi{} is split temporally into a train, validation, and test set.
    \hifi{} is evenly spread in time, with quarterly peaks.}
    \label{fig:temporalsplit}
\end{figure}

\paragraph{Temporal Split.}
We split the dataset temporally to eliminate forward-looking bias, so the model is not able to infer previous extractions based on things learned from the future. 
The temporal split means we have a short time window for the test and validation split.
Therefore, most companies only report a 10-K in either the validation split or the test split timeframe.
Therefore, we make the validation split more representative by using company-specific cutoff dates. 
We still follow a temporal split, setting the company-specific cutoff date, resulting in something as close to a 50/50 split as possible.
If only one filing is present, the choice is random. 
This approach leads to 89.88\% of companies in the validation set also appearing in the test set.
Finally, all companies with their first filing after 2023-10-31 are assigned to the test set to better evaluate generalization to previously completely unseen domains.
The final temporal split of entries in the dataset can be seen in Figure \ref{fig:temporalsplit}.

\begin{table*}[t]
\centering
\resizebox{\textwidth}{!}{
\begin{tabular}{@{}lrrrrrrrr@{}}
\toprule
\textbf{Error Category} & \multicolumn{2}{c}{\textbf{Train}} & \multicolumn{2}{c}{\textbf{Validation}} & \multicolumn{2}{c}{\textbf{Test}} & \multicolumn{2}{c}{\textbf{Total}} \\
\cmidrule(lr){2-3} \cmidrule(lr){4-5} \cmidrule(lr){6-7} \cmidrule(lr){8-9}
& Before & After & Before & After & Before & After & Before & After \\
\midrule
\textit{Sample} & 277 (100) & 287 (100) & 280 (100) & 285 (100) & 257 (100) & 242 (100) & 814 (300) & 814 (300) \\
\midrule
\multicolumn{9}{@{}l}{{Errors -  entity count (entry count)}} \\
\addlinespace[2pt]
\quad Date     & 1 (1) & 0 (0) & 2 (2) & 0 (0) & 0 (0) & 1 (1) & 3 (3)   & 1 (1) \\
\quad Currency & 7 (7) & 0 (0) & 7 (4) & 0 (0) & 8 (6) & 0 (0) & 22 (17) & 0 (0) \\
\quad Span     & 5 (4) & 2 (2) & 5 (4) & 3 (3) & 10 (5)& 0 (0) & 20 (13) & 5 (5) \\
\quad Value    & 0 (0) & 1 (1) & 0 (0) & 0 (0) & 0 (0) & 0 (0) & 0 (0)   & 1 (1) \\
\midrule
\multicolumn{9}{@{}l}{{Parsing Artifacts (entry count)}} \\
\addlinespace[2pt]
\quad Malformed  & - (11)    & - (2)     & - (3)     & - (1)     & - (8)     & - (4)     & - (22)     & - (7) \\
\midrule
Errors & 13 ({23}) &  3 (5) & 14 (13) & 3 (4) & 18 (19) & 1 (5) & 45 (55) & 7 (14) \\
\midrule
Error Rate (\%) & 4.7 (23.0) & 1.0 (5.0) & 5.0 (13.0) & 3.7 (4.0) & 7.0 (19.0) & 0.4 (5.0) & 5.5 (18.3) & 0.8 (9.3) \\
\bottomrule
\end{tabular}
} 
\caption{\textbf{Error analysis of \hifi{}.} 
The table compares error counts from a manual audit of 300 random entries (100 per split) \textit{before} and \textit{after} applying a regex-based cleaning filter. 
We report both the entity-level metrics and entry-level metrics in parentheses. 
The final entity-level error rate is less than 1\%.
}
\label{tab:error_analysis_comparison}
\end{table*}

\paragraph{Taxonomy Creation.}

\begin{figure*}[t]
\centering  
\includegraphics[width=\linewidth]{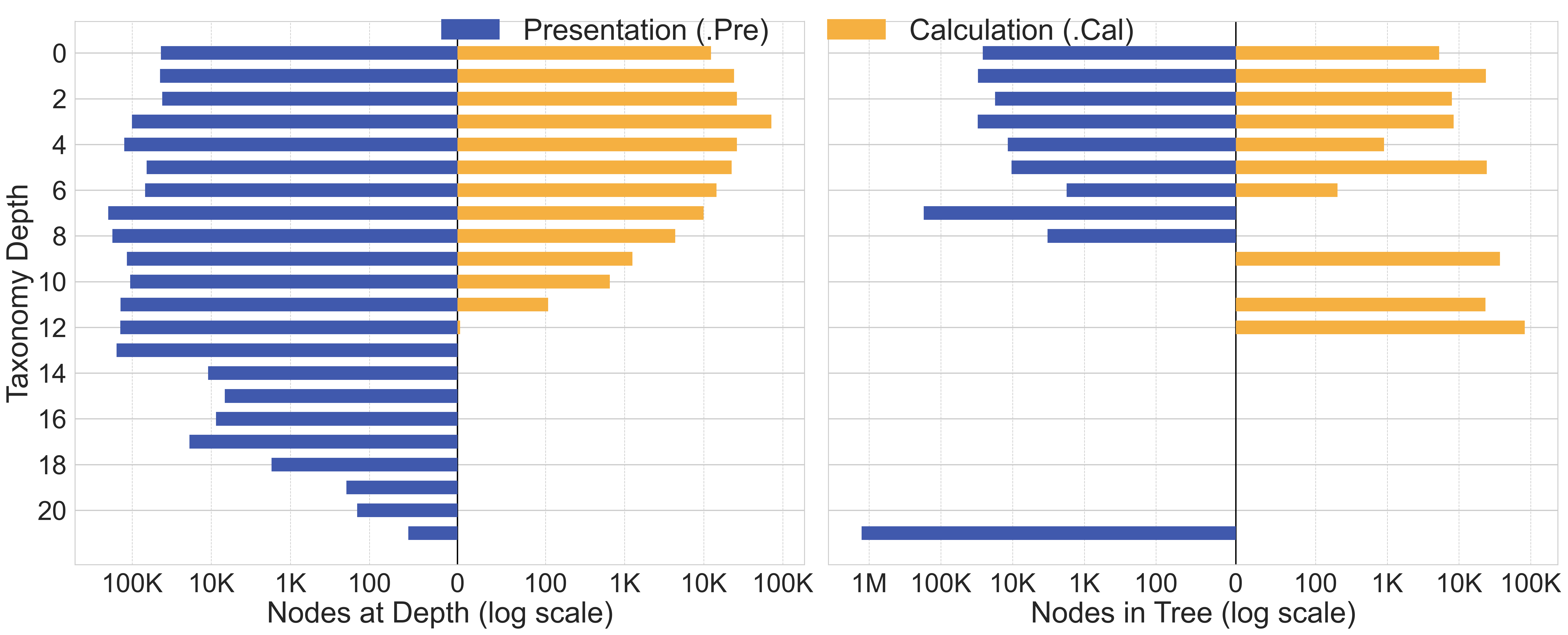}
\caption{\textbf{Taxonomy Statistics} The plot shows how many nodes are at a given depth (left), and the size of a tree at a given depth (right). 
The Presentation taxonomy is deeper, though both taxonomies distribute nodes without pronounced clustering at any particular level.}
\label{fig:taxonomy_comparison}
\end{figure*}

Each iXBRL report has several file attachments describing parent--child relationships. 
We focus on the relationships in two of the complex formats: \texttt{.cal} and \texttt{.pre}.
We download attachments and parse them with Arelle \citep{Arelle_GitHub} to JSON.
Since different companies use the taxonomy in different ways, children can have multiple parents.
Therefore, we aggregate the per-document hierarchies to build two unified taxonomies.
We do this by the following formula, using the most common parent as the parent. $P_{\text{master}}$: 
\[
P_{\text{master}}(t) = \arg\max_{p \in \mathcal{P}(t)} \text{count}(p, t),
\]
where \(t\) is a tag, \(\mathcal{P}(t)\) its possible parents, and \(\text{count}(p, t)\) the frequency of documents with the parent--child relation. In the very rare case of ties, we pick randomly.
Figure~\ref{fig:taxonomy_comparison} shows statistics for the unified Presentation (1.57M edges, depth=21) and Calculation (202K edges, depth=12) taxonomies.
Figure~\ref{fig:taxonomy_comparison} shows that the Calculation and Presentation taxonomies both provide an even spread of nodes across all depths in the taxonomy, of course, with fewer nodes at the very top of the tree.
Further Figure~\ref{fig:taxonomy_comparison} also shows how the Presentation taxonomy allows for significantly more depth and nuance in its layers, with a large percentage of nodes belonging to a tree that goes all the way to depth 21, compared to the significantly less deep Calculation taxonomy.

\paragraph{Bottom-up Aggregation Algorithm.}
\begin{algorithm}[h!]
\DontPrintSemicolon
\KwIn{
  \begin{itemize}
  \setlength\itemsep{0em}
    \item A hierarchical taxonomy $T$, 
    \item Number of collapse steps $n$ 
  \end{itemize}
}

\textbf{Iterative Collapsing.}\\
    \For{$i = 1$ to $n$}{
        \begin{enumerate}
          \item Let $L$ be the set of all leaf nodes in $T$. 
          \item \ForEach{leaf $l \in L$}{
            Given $l$ is not a root\\
            Let $p$ be the parent of $l$.\\
            Replace $l$ with $p$ in $T$. \\
            
          }
        \end{enumerate}
    }
\Return{$T$}
\caption{Taxonomy-Based Grouping via Bottom-Up Selection}
\label{alg:taxonomy_collapse}
\end{algorithm}

We employ a bottom-up hierarchical aggregation method to address the sparsity of our initial $\sim$ 200,000 fine-grained labels. 
Starting from the leaf nodes, we iteratively map each label to its immediate parent, which systematically reduces specificity while increasing the cardinality of each class. 
This bottom-up approach is particularly effective as our taxonomy is narrow at the uppermost layers, meaning a top-down approach would change nothing for a lot of labels.
The procedure is detailed in Algorithm~\ref{alg:taxonomy_collapse}. 

\paragraph{\hifi{}-Lite.}
\begin{table*}[t]
    \centering
    \scriptsize 
    \setlength{\tabcolsep}{4pt} 
    \begin{tabularx}{\textwidth}{>{\raggedright\arraybackslash}X r}
        \toprule
        \textbf{Label (\hifi{})} & \textbf{Concept (\hifi-Lite)} \\
        \midrule
        us-gaap:IncomeLossFromContinuingOperations & Earnings \\
        us-gaap:NetIncomeLoss & Earnings \\
        ... \\
        \midrule
        us-gaap:OperatingIncomeLoss & EBIT \\
        \midrule        cmtl:WeightedAveragePerformanceSharesOutstandingDuringThePeriodThatAreExcludedfromEPSCalculation & EPS \\
        enb:WeightedAverageInterestInOwnCommonShares & EPS \\
        ... \\
        \midrule
        us-gaap:FeeIncome & Revenues \\
        us-gaap:InsuranceCommissionsAndFees & Revenues \\
        .. \\
        \bottomrule
    \end{tabularx}
    \caption{\textbf{Industry Expert Mappings Excerpt.} \hifi{}- is made by using mappings selected by an industry expert to generalizable financial Concepts. 
    Two of each category were selected. 
    The full set is available in appendix \ref{ExpertMappings}.
    }
    \label{tab:label_mapping}
    
\end{table*}
We collaborated with a senior domain expert with over a decade of experience, leading a department at a top quantitative finance firm. 
We investigate the possibilities of a dataset based on manual concept linking.
To create \hifi{}-Lite, we mapped selected iXBRL terms to their corresponding general finance concepts as shown in Table~\ref{tab:label_mapping}.  
The financial expert selected four key financial figures for evaluating financial reports: \emph{Earnings}, \emph{EBIT}, \emph{EPS}, and \emph{Revenues}, and then identified relevant iXBRL tags.
\hifi{}-Lite is created from \hifi{} by applying the expert mapping to convert XBRL labels. 
To make a very curated and relevant subset, we only retain snippets with more than half the entities being part of the expert mappings.
\section{Experiments}
We demonstrate the usefulness of our context-rich dataset and granularity selection method by establishing three baselines.
Text classification, Sequence labeling, and structured information extraction with LLMs. 
To better describe the performance of our models on the large label set.
We report the macro-F1 score as a function of cumulative support, where tags are incrementally included in the calculation, ordered from most to least frequent.
(Details about metric calculations in Appendix~\ref{app:calculation}.)

\paragraph{Text Classification.}
We define a simple classification task: Predict the first entity's label from the entry. 
We embed each entry with \texttt{EmbeddingGemma-300M}~\citep{choi2025introducing} and then fine-tune 
only the classification head using the Adam optimizer~\citep{kingma2017adammethodstochasticoptimization}  with a learning rate of $1\times10^{-5}$ for 20 epochs on the training set.
We do this for iterations $n=\{1-10\}$ of our bottom up selection Algorithm~\ref{alg:taxonomy_collapse}.

\begin{figure*}[t]
    \centering
    \begin{minipage}[c]{0.03\linewidth}
        \centering
        \rotatebox{90}{\scriptsize Aggregate Macro F$_1$}
    \end{minipage}
    \begin{minipage}[c]{0.95\linewidth}
        \includegraphics[width=0.32\linewidth]{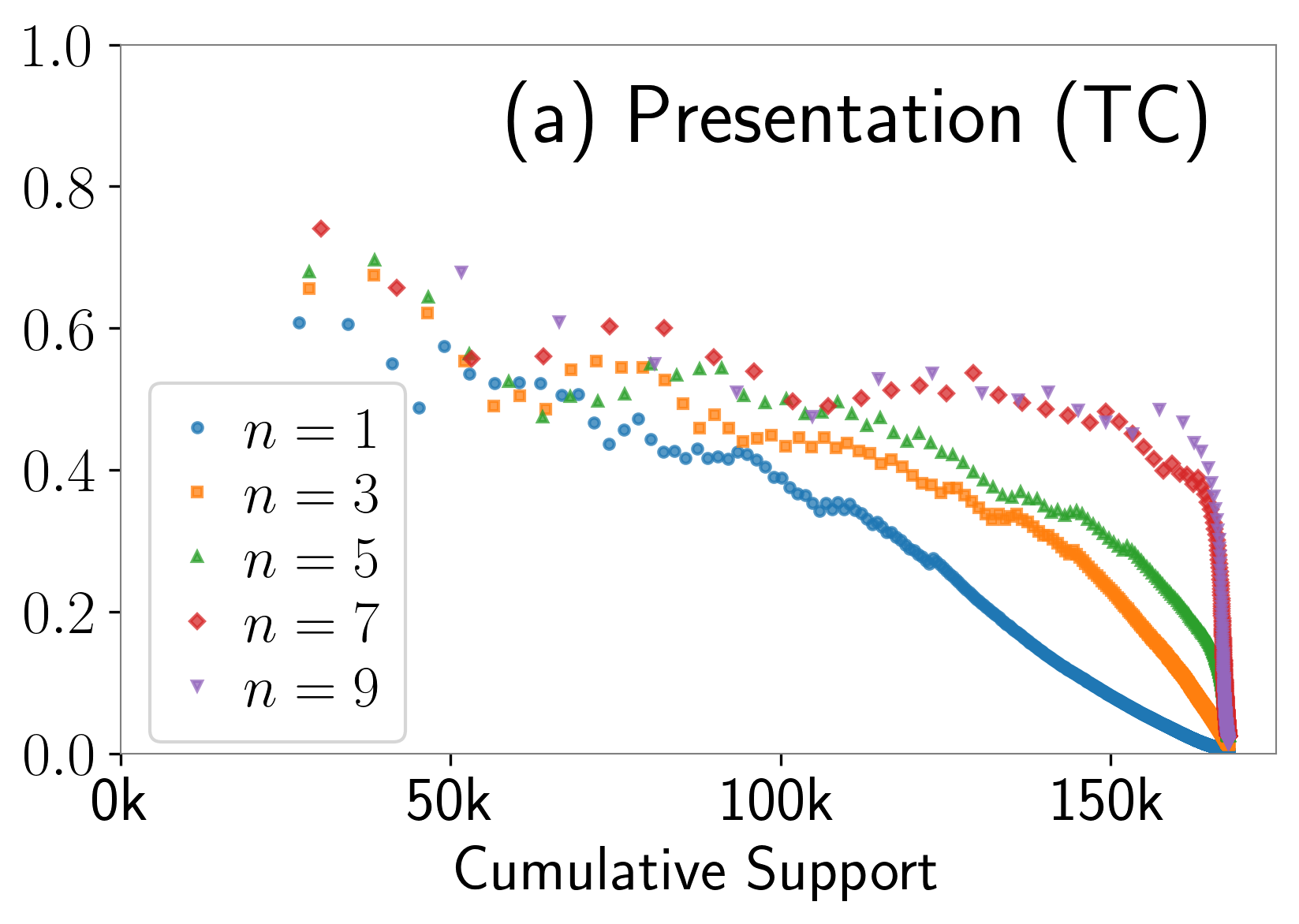}\hfill
        \includegraphics[width=0.32\linewidth]{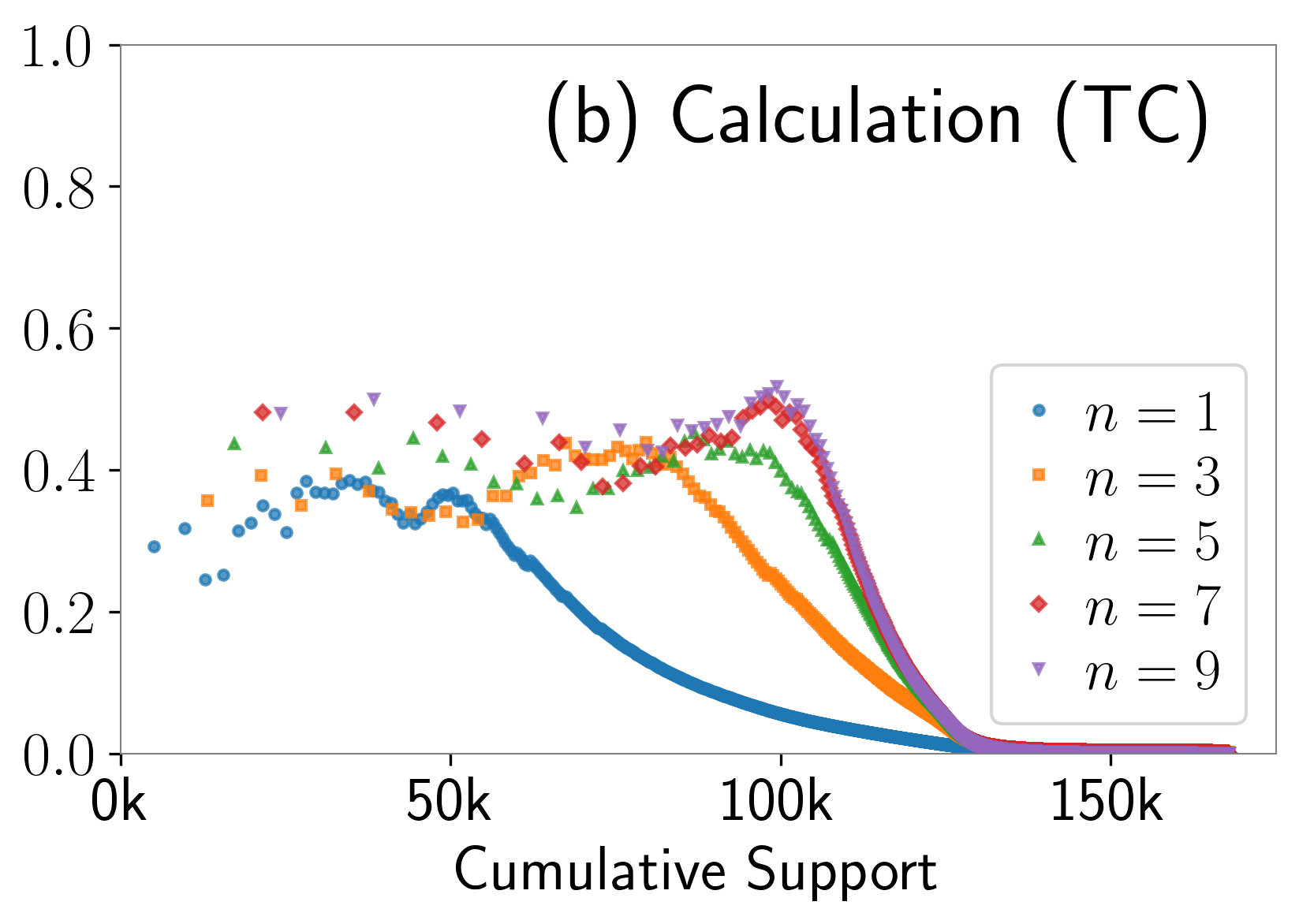}\hfill
        \includegraphics[width=0.32\linewidth]{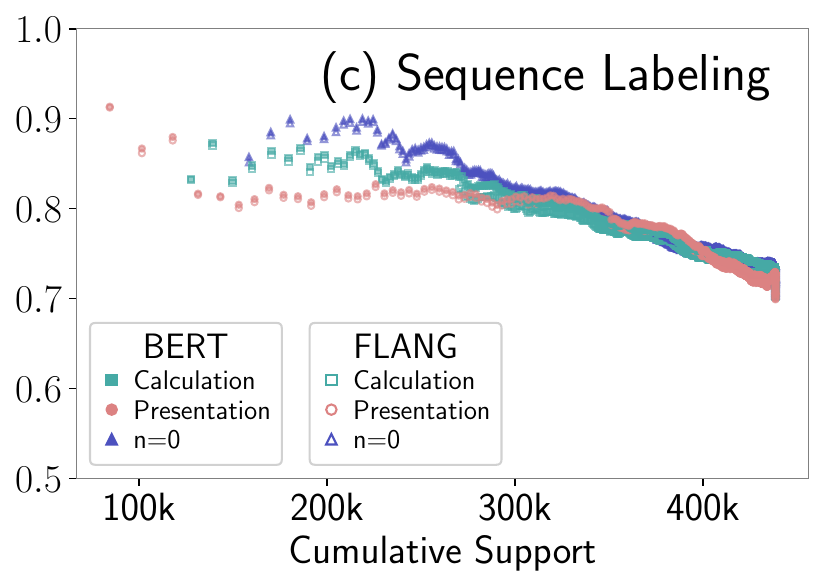}
    \end{minipage}
    
    \caption{\textbf{Results HiFi-KPI.} Plots show aggregate macro-F$_1$ scores as a function of cumulative label support, on the labelset derived from n iterations of our bottom-up method. 
    Labels are ordered from the most to the least common tag (left to right). 
    Models achieve strong performance on well-represented labels across all levels of granularity, with F$_1$ scores decreasing for less frequent labels. 
    Subplots correspond to: Presentation taxonomy (left), Calculation taxonomy (middle), and Sequence Labeling task (right).}
    \label{fig:layer-comparison}
\end{figure*}

\paragraph{Sequence Labeling.}
The task is to identify and classify each token with a label. 
To establish strong initial baselines while managing computational resources, the sequence labeling experiments focus on the 1,000 most frequent tags for a given label set.
Tags outside this set are mapped to a \textit{OOS} (out-of-scope) label.
We use \texttt{bert-base-uncased}~\citep{devlin2019bertpretrainingdeepbidirectional} and \texttt{FLANG-BERT}~\citep{shah-etal-2022-flue}, a BERT model adapted for the financial domain, with a standard token classification head, we fine-tune full models using the Adam optimizer~\citep{kingma2017adammethodstochasticoptimization}, a learning rate of $1\times10^{-5}$ and max.\ 50 epochs with early stopping patience of 2. 
We show the utility of the taxonomy, while being mindful of compute resources, by training a model on the $n=1$ of the \texttt{.cal} and \texttt{.pre} taxonomies.

\paragraph{LLM-Based Structured Data Extraction.}
Finally, we use \hifi{}-Lite to evaluate LLM-based structured extraction, including tags, dates, currency, and numeric values. We compare four LLMs: Gemma-3-27B~\cite{gemmateam2025gemma3technicalreport}, Qwen3-30B-A3B~\citep{qwen3technicalreport} and mistral-Small-3.2-24B~\citep{mistral2025small}, DeepSeek-V3.1~\cite{deepseekai2024deepseekv3technicalreport}.
We use a 1-shot prompt, where we describe how we want to extract the data and give an example of a single extraction(Full prompt avalible in appendix \ref{app:system}). 
We evaluate the models on how well they extract fully identical entities with the gold standard. 
Further, we evaluate it as a label extraction task, where the goal is to extract the correct labels independent of the contextual information.
We do 5 runs of each LLM on the Lite set, except for DeepSeek-V3.1.

\begin{table*}[t]
\centering
\small
\begin{tabular}{llllll}
\toprule
& \multicolumn{4}{c}{\textbf{Full Entity Extraction}} & \multicolumn{1}{c}{\textbf{Label Extraction}} \\
\cmidrule(lr){2-5} \cmidrule(lr){6-6}
\textbf{Model} & {\textbf{F1}} & {\textbf{Precision}} & {\textbf{Recall}} & {\textbf{Jaccard}} & {\textbf{Macro-F1}} \\
\midrule
BERT (SL)                 & {-} & {-} & {-} & {-} & \bfseries 0.914 \\
FLANG-BERT (SL)           & {-} & {-} & {-} & {-} & 0.907 \\
EmbeddingGemma-300m (TC)  & {-} & {-} & {-} & {-} & 0.521\\
\midrule
Mistral-small-3.2-24B     & 0.422$_{\pm0.008}$ & 0.423$_{\pm0.008}$ & 0.422$_{\pm0.008}$ & 0.268$_{\pm0.006}$ & 0.470$_{\pm0.015}$ \\
Gemma-3-27B-it            & 0.356$_{\pm0.008}$ & 0.358$_{\pm0.008}$ & 0.355$_{\pm0.008}$ & 0.217$_{\pm0.006}$ & 0.543$_{\pm0.030}$ \\
Qwen3-30B-A3B             & \textbf{0.440}$_{\pm0.008}$ & 0.448$_{\pm0.008}$ & 0.433$_{\pm0.009}$ & 0.282$_{\pm0.007}$ & 0.543$_{\pm0.005}$ \\
DeepSeek-V3.1             & {0.436} & {0.463} & {0.413} & {0.279} & 0.464 \\
\bottomrule
\end{tabular}
\caption{\textbf{Results \hifi{}-Lite}, We report the F1, Precision, Recall, and Jaccard similarity as well as the SD based on 5 runs, except for DeepSeek-V3.1. 
We report their match with gold entities, as well as their performance on the label extraction task.
The encoder-based text classification sequence labelling models can not extract full entities.
The table shows the best performance by Qwen3 on entity extraction and BERT (Sequence Labelling) for the label extraction.}
\label{tab:comparison}
\end{table*}

\section{Results and Analysis}

\subsection{Quantitative Results and Analysis}

\paragraph{Text Classification.}
Figure~\ref{fig:layer-comparison} shows that more coarse-grained labels of the \texttt{.pre} taxonomy generally boosts macro-F1, as the bottom-up approach reduces label sparsity. 
For \texttt{.cal}, performance saturates quickly, possibly because its hierarchy is less deep. 
The model struggles with infrequent tags, showing opportunities to develop stronger methods.
\paragraph{Sequence Labeling.}
Sequence labeling shows higher performance than text classification.
We observe the effectiveness of the \texttt{.pre} taxonomy, as the most common label the special (out-of-scope) label has significantly lower support than in the Calculation or the ungrouped. 
The most common ungrouped tags achieve the highest macro F1, followed by the most common in the \texttt{.cal}, and then \texttt{.pre}, before all representations converge to the same macro F1 for long-tail labels.
We observe that the difference between the FLANG and BERT-based models is close to non-existent.
The performance for this task is high, especially considering the 1000 different labels.

\paragraph{LLM-based Extraction.}
Table~\ref{tab:comparison} shows very consistent performance across models.
Overall, performance demonstrates the difficulty of extracting full, well-structured records from financial text.
One common theme is failures in normalizing values correctly to match the gold standard.
Interestingly, DeepSeek-V3.1 demonstrates a more conservative extraction strategy, achieving the highest precision among all models at the expense of lower recall.
opposite of Qwen that seems to have the greatest tendency to predict a label even with low confidence, instead of using the (out-of-scope) token.
However, most errors performed by the LLMs seem to be inconsistent and have variety in their form. 

\subsection{Qualitative Analysis}
\paragraph{Text Classification.}
Inspecting the classification results, the model predictions have granularity errors, for example, the second most common error for both $n=1$ and $n=3$ is respectively \texttt{\seqsplit{us-gaap:AccountsNotesAndLoansReceivableLineItems}} predicted as \texttt{\seqsplit{us-gaap:DebtInstrumentLineItems}} and \texttt{\seqsplit{us-gaap:StatementOfCashFlowsAbstract}} as \texttt{\seqsplit{us-gaap:BusinessAcquisitionLineItems}}.
Both cases highlight that, likely due to more training samples, the model has a tendency to favor the more common label.
This is consistent with Figure \ref{fig:layer-comparison}, which shows that higher support correlates with higher performance.
This highlights the potential of future work investigating hierarchical classification methods like \citet{zhou2020hierarchy, agrawal2025hierarchical, jain2024higen}

\paragraph{Sequence Labeling.}
Manual review shows that both FLANG and BERT-based models have close to perfect performance on the task. 
An interesting pattern is, for example, in the 10-K from First BanCorp on December 31, 2023. 
\begin{quote}
    As of December 31, 2023, the Company's securities portfolio held 657 securities of which 632 securities were in an unrealized loss position. 
\end{quote}
The long-tailed nature of the data results in neither the calculation nor the $n=0$ model having the tag \texttt{\seqsplit{fbnc:DebtSecuritiesAvailableForSaleAndHeldToMaturityNumberOfPositions}} because it falls outside the 1000 most common tags, whereas the \texttt{.pre} based model uses the more general tag \texttt{\seqsplit{us-gaap:ScheduleOfTradingSecuritiesAndOtherTradingAssetsLineItems}}, correctly classifying with higher abstraction and robustness. 
The $n=0$ and \texttt{.cal} based model correctly realizes that this tag is outside their dataset's label set.
This correct realization shows potential for a future system that conditionally uses the different models.
This again highlights the flexibility of the \texttt{.pre} compared to the \texttt{.cal}. 
This also shows that \hifi{} can be used to successfully train models that use these hierarchies to go from a more specific tag with lower confidence to a more abstract tag with higher confidence and robustness. 
To answer \textbf{RQ1}, our qualitative and quantitative analysis shows how \hifi{} can be used to correctly classify a larger portion of the dataset compared to using more specific tags.
Further for \textbf{RQ1.1}, we find that the \texttt{.pre} provides the most flexible representation and best enables training of models for a specific level of granularity.
This is likely because the \texttt{.pre} is deeper than the \texttt{.cal} and has a more varied distribution of node depths, as illustrated in Figure~\ref{fig:taxonomy_comparison}. 

\paragraph{LLM-based Extraction}
For the LLM-based extraction on \hifi{}-Lite, one theme is that the \texttt{Gemma-27B} does not understand "three months ended". (Example in Figure~\ref{fig:gemma-date-error}.)

\definecolor{gold}{RGB}{212, 175, 55}

\begin{figure}[h!]
    \centering 
    \textbf{Text}
    \begin{quote}
        \textit{"The Reciprocal Exchanges generated \$61 million of earned premiums for the three months ended March 31, 2024."}
    \end{quote}
    \small\textbf{\textcolor{gold}{Gold Standard}}
    \begin{lstlisting}[style=jsonstyle]
{ 'label': 'revenues',
  'start_date_for_period': '2024-01-01',
  'end_date_for_period': '2024-03-31',
  'currency_unit': 'USD',
  'value': 61000000.0 }    \end{lstlisting}

    \small\textbf{Gemma-27B Prediction}
    \begin{lstlisting}[style=jsonstyle]
{ 'label': 'revenues',
  'start_date_for_period': '@@2024-03-31@@',
  'end_date_for_period': '2024-03-31',
  'currency_unit': 'USD',
  'value': 61000000.0 }    \end{lstlisting}
\caption{{Gemma-27B} incorrectly predicts start dates, using the same \texttt{start\_date\_for\_period} as \texttt{end\_date\_for\_period}. 
All other models predict the right \textcolor{gold}{Gold} date.}
    \label{fig:gemma-date-error}
\end{figure}

\noindent \texttt{DeepSeek-V3.1} is less prone to extraction, clear from its lower recall but highest of all precision.
Most LLM extraction errors are inconsistent and have more variety. 
One example of errors is that the Qwen model has a greater tendency than other models to guess on a label instead of using the (out-of-scope) label.
One other consistent thing is that the models struggle to normalize values correctly to match the gold standard.
Lastly, it is important to note the quite strict criteria for this task, which requires it to extract a totally correct entity, especially considering the simple rules \hifi{}-Lite is created from, where multiple interpretations can be correct.

To answer \textbf{RQ1.2}, we find that state-of-the-art LLMs show promise on \hifi{}-Lite; however, due to the very specific domain and complexity of the task, they struggle with accurately performing the specified task, especially in understanding magnitude classifiers, domain-specific terms like "basis points".
Finally, our fine-tuned embedding models, especially for sequence labeling, outperform LLM-based models at label classification, which again highlights the value of domain-specific datasets.
\hifi{} and our initial models represent a significant step towards the automatic extraction of financial KPIs and iXBRL tagging. 
For investors, this automation fulfills the original promise of iXBRL by delivering truly structured and analyzable data. 
This helps prevent major investment mistakes based on flawed premises, such as an incorrect date being used for a tag. 
Finally, the system equips legal and regulatory bodies with a more efficient tool for verifying compliance.

\section{Conclusion}
In this paper, we introduce \hifi{} consisting of 1.65M paragraphs with 4.5M annotations derived from SEC filings.
\hifi{} takes a significant step towards building an automated system for KPI tag extraction, by creating a more generalizable financial NLP resource than previously available based on financial reports. 
\hifi{} has two unified taxonomies \texttt{.pre} and \texttt{.cal} that structure the iXBRL label set, making it possible to select a specified granularity. 
Further \hifi{} is the first set to provide valuable contextual details for labels, including temporal, currency, and numeric values. 
We report initial baselines for this new resource with encoder-based classification models as well as LLM-based structured extraction of these new contextual details. 
Finally, to facilitate rapid prototyping and evaluation, as well as hint at what is possible, we also introduce \hifi{}-Lite with expert mappings, showing the potential for even better aggregation algorithms.
Our analysis shows that fine-tuned encoder-based models achieve strong performance on the label extraction task, while SOTA LLMs show potential with entity extraction from \hifi{}-Lite, further improvement is certainly possible.

\section*{Limitations}
A limitation of our experiments is that the annotation quality may vary throughout our dataset, evidenced by the fact that the SEC regularly publishes Data Quality Reminders. 
For instance, the SEC has noted that some filers use different labels for the same element on income statements across periods \cite{sec_changing_labels_2023} or don't report the most fundamental key figure, earnings per share, correctly \cite{sec_eps_tagging_2024}.
Lastly, there is a bias in the dataset created as the text snippets in the dataset consist only of the spans in these documents, and we only include snippets that match our simple parser methodology.
\section*{Ethical Considerations}
\hifi{} is built from public 10-K and 10-Q reports filed with the U.S. Securities and Exchange Commission (SEC) intended for public disclosure. 
Our scraping methodology adheres to the guidelines and terms laid out by the SEC.

The dataset is subject to two potential sources of bias. 
The bias towards bigger companies, as they are more likely to be publicly traded, and required to file with the SEC. 
Second, a geographical bias towards the US. 
Therefore, findings may not generalize to companies operating under different international financial reporting standards.

All human labor involved in the creation of the dataset was compensated fairly. 
The senior domain expert who assisted in the creation of \hifi{}-Lite was compensated at his usual hourly rate. 
The manual data quality checks were also performed by a compensated professional.

Training large-scale models can have a significant environmental impact. 
To mitigate this, we have limited our experiments to what is necessary for establishing strong baselines. Further, to address this, we create the \hifi{}-Lite subset. 
This smaller, curated resource allows for rapid and efficient model evaluation, significantly lowering the barrier to entry for researchers and reducing the overall environmental impact of working with our data.

\section*{Acknowledgments}
We would like to thank the AAU-NLP group for helpful discussions and feedback on an earlier version of this article. 
We want to also thank Alipes ApS for their support in facilitating and funding this research and the useful discussions with their Quant NLP team.
Rasmus Aavang is supported by the Industrial Ph.D. programme from Innovation Fund Denmark (grant code 4297-00016B).
MZ and JB, were supported by the research grant (VIL57392) from VILLUM FONDEN. 
MZ also received funding from the Danish Government to Danish Foundation Models (4378-00001B).

\section*{Bibliographical References}

\bibliographystyle{lrec2026-natbib}
\bibliography{lrec2026-example}

\section{Language Resource References}
\label{lr:ref}
\bibliographystylelanguageresource{lrec2026-natbib}
\bibliographylanguageresource{languageresource}

\clearpage
\appendix

\section{Elaboration on Metric Calculation, Defining Precision, Recall, Micro F1 and Macro F1}
\label{app:calculation}
For the \hifi{} Lite set, we define precision, recall, micro F1, and macro F1 using an adapted approach, as generative LLM predictions are unrestricted. 
A misclassified prediction is counted as a false negative for the true label and a false positive for the predicted label. 
A correct prediction is counted as a true positive. 
Using these definitions, we compute micro F1 as in any standard classification task.
For macro F1, we take the average F1 score of only the ground truth labels, excluding labels that appear solely in the predicted set.

For the Figure \ref{fig:layer-comparison}, we compute the cumulative sum by iterating over the label distribution from the most frequent to the least frequent label in the test set. 
We then calculate the macro-average F1 score for the top x included labels.
\clearpage
\onecolumn
\section{System prompt}\nopagebreak
\label{app:system}
\begin{tcolorbox}[title=System Prompt, promptstyle]
\lstset{
    basicstyle=\normalfont\sffamily\tiny,
    breaklines=true,
    frame=none,
    columns=fullflexible,
}
\begin{lstlisting}[linewidth=\linewidth]
###################
### System Prompt ###
###################

You are an expert data extraction assistant. Your task is to read a given text and extract financial or entity-related information. For each entity found in the text, extract:

        - **value**: numerical representation;
        - **currency_/_unit**: currency most often USD, shares, EUR, CAD, etc.;
        - **label**: revenues, earnings, eps, ebit, or XBRL-OOS (if it is none of the others);
        - **start_date_for_period**: (if available)
        - **end_date_for_period**: (if available)


        If no relevant data is found, return an empty list. Otherwise, return a json line of one or more dictionaries after the "entities" key, each containing these fields exactly:

        [
            {
                "entities": [
                    {
                    "label": "<extracted label>",
                    "start_date_for_period": "<YYYY-MM-DD>",
                    "end_date_for_period": "<YYYY-MM-DD>",
                    "currency_/_unit": "<unit or currency>",
                    "value": <numeric value>
                    }
                ]
            }
        ]

        No additional commentary or text should be included, only valid JSON.

        Example:

        Text: The Company has incurred losses since 2008 resulting from a combination of: declining net interest income, as our loan portfolio decreased from $109.8 million at December 31, 2008 to $62.3 million at December 31, 2016; increased provisions for loan losses between 2009 and 2012; and increasing non-interest expense related to professional fees and repossessed asset write-downs and costs. The Company recently incurred net losses of $866 for the nine months ended September 30, 2017 and $1,260 during the year ended December 31, 2016.  Our interest income for the nine months ended September 30, 2017 has increased with the increase in the balance of our loan portfolio, however, this growth has also resulted in an increase to our provision for loan losses.  Our non-interest expense has also increased for compensation and occupancy cost and includes costs for problem asset resolution at the beginning of the year. The loss for 2016 was largely a result of our net interest income reflecting the low balance of our loan portfolio, increasing professional fees for problem asset resolution and additional costs associated with operating as a public company. Non-interest expense for 2016 was also impacted by an operational loss not reimbursable from our insurance.
        
        [
            {
                "entities": [
                    {
                        "label": "XBRL-OOS",
                        "start_date_for_period": "2008-12-31",
                        "end_date_for_period": "2008-12-31",
                        "currency_/_unit": "USD",
                        "value": 109800000.0
                    },
                    {
                        "label": "XBRL-OOS",
                        "start_date_for_period": "2016-12-31",
                        "end_date_for_period": "2016-12-31",
                        "currency_/_unit": "USD",
                        "value": 62300000.0
                    },
                    {
                        "label": "earnings",
                        "start_date_for_period": "2017-01-01",
                        "end_date_for_period": "2017-09-30",
                        "currency_/_unit": "USD",
                        "value": -866000.0
                    },
                    {
                        "label": "earnings",
                        "start_date_for_period": "2016-01-01",
                        "end_date_for_period": "2016-12-31",
                        "currency_/_unit": "USD",
                        "value": -1260000.0
                    }
                ]
            }
        ]

\end{lstlisting}
\end{tcolorbox}
\twocolumn
\twocolumn[
\section{Finance Expert Handpicked Labels and Their Meaning}
\label{ExpertMappings}
\vspace{1em} 
]
\begin{table}[]
    \centering
    \small
    \resizebox{\textwidth}{!}{ 
    \begin{tabular}{ll}
        \toprule
        \textbf{Label} & \textbf{Category} \\
        \midrule
        us-gaap:IncomeLossAttributableToParent & Earnings \\
        us-gaap:IncomeLossFromContinuingOperations & Earnings \\
        us-gaap:IncomeLossFromContinuingOperationsBeforeIncomeTaxesExtraordinaryItemsNoncontrollingInterest & Earnings \\
        us-gaap:IncomeLossFromContinuingOperationsBeforeIncomeTaxesMinorityInterestAndIncomeLossFromEquityMethodInvestments & Earnings \\
        us-gaap:NetIncomeLoss & Earnings \\
        us-gaap:NetIncomeLossAvailableToCommonStockholdersBasic & Earnings \\
        us-gaap:OperatingIncomeLoss & EBIT \\
        bw:IncrementalCommonSharesAttributableToDilutiveEffectOfNetIncome & EPS \\
        cmtl:WeightedAveragePerformanceSharesOutstandingDuringThePeriodThatAreExcludedfromEPSCalculation & EPS \\
        enb:WeightedAverageInterestInOwnCommonShares & EPS \\
        fcx:DilutiveSecuritiesExcludedfromComputationofEPSAmount & EPS \\
        gpmt:AntidilutiveSecuritiesExcludedfromComputationofEarningsPerShareInterestExpense & EPS \\
        gs:ImpactOfUnvestedShareBasedPaymentAwardsAsSeparateClassOfSecuritiesOnEarningsPerShareBasic & EPS \\
        land:WeightedAverageNumberOfOperatingPartnershipUnitsHeldByNoncontrollingInterest & EPS \\
        pcg:PlanOfReorganizationBackstopCommitmentPremiumCommonStockShares & EPS \\
        us-gaap:DistributedEarnings & EPS \\
        us-gaap:DividendsAndInterestPaid & EPS \\
        us-gaap:EarningsPerShareBasic & EPS \\
        us-gaap:EarningsPerShareBasicAndDiluted & EPS \\
        us-gaap:IncrementalCommonSharesAttributableToConversionOfDebtSecurities & EPS \\
        us-gaap:IncrementalCommonSharesAttributableToParticipatingNonvestedSharesWithNonForfeitableDividendRights & EPS \\
        us-gaap:IncrementalCommonSharesAttributableToShareBasedPaymentArrangements & EPS \\
        us-gaap:ParticipatingSecuritiesDistributedAndUndistributedEarningsLossBasic & EPS \\
        us-gaap:UndistributedEarnings & EPS \\
        us-gaap:WeightedAverageNumberOfSharesContingentlyIssuable & EPS \\
        us-gaap:WeightedAverageNumberOfSharesRestrictedStock & EPS \\
        us-gaap:DirectFinancingLeaseRevenue & Revenues \\
        us-gaap:FeeIncome & Revenues \\
        us-gaap:InsuranceCommissionsAndFees & Revenues \\
        us-gaap:OperatingLeaseLeaseIncome & Revenues \\
        us-gaap:PremiumsEarnedNet & Revenues \\
        us-gaap:Revenues & Revenues \\
        us-gaap:UnregulatedOperatingRevenue & Revenues \\
        us-gaap-supplement:FeeIncome & Revenues \\
        us-gaap-supplement:InterestIncomeOperatingPaidInKind & Revenues \\
        \bottomrule
    \end{tabular}
    }
    \caption{Mapping of XBRL labels to expert labels.}
    \label{tab:label_mapping}
\end{table}
\end{document}